%% file: 0_main.tex
\pdfoutput=1

\documentclass[11pt]{article}

\usepackage[preprint]{acl}

\usepackage{times}
\usepackage{latexsym}
\usepackage{amsmath}
\usepackage{booktabs}
\usepackage{fontawesome}

\usepackage[T1]{fontenc}

\usepackage[utf8]{inputenc}

\usepackage{microtype}

\usepackage{inconsolata}
\usepackage{caption}
\usepackage{stfloats}
\usepackage{xspace}

\usepackage{graphicx}
\usepackage{tabularx}
\usepackage{tcolorbox}
\usepackage{adjustbox}
\usepackage{bm}
\usepackage{subfigure}
\usepackage{soul}
\usepackage{color}
\usepackage{comment}
\usepackage{colortbl}
\usepackage{multirow}

\newcommand{\copilotlogo}[1][1.4em]{%
  \raisebox{-0.3\height}{\includegraphics[height=#1]{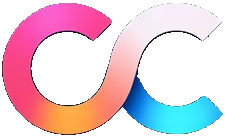}}
}
\newcommand{\ours}{ComfyUI-Copilot\xspace}

\title{\copilotlogo\hspace{3pt}ComfyUI-Copilot: An Intelligent Assistant for Automated Workflow Development}

\author{
  Zhenran Xu\textsuperscript{1,2}\thanks{Work done during internship at Alibaba International Digital Commerce.}, Xue Yang\textsuperscript{1}\thanks{Corresponding authors.}, Yiyu Wang\textsuperscript{1}, Qingli Hu\textsuperscript{1}, Zijiao Wu\textsuperscript{1}, \\
  \textbf{Longyue Wang\textsuperscript{1}, Weihua Luo\textsuperscript{1}, Kaifu Zhang\textsuperscript{1}, Baotian Hu\textsuperscript{2}\footnotemark[\value{footnote}], Min Zhang\textsuperscript{2}} \\
  \textsuperscript{1}Alibaba International Digital Commerce \quad \textsuperscript{2}Harbin Institute of Technology (Shenzhen)\\
  \texttt{\{xuzhenran.xzr,shali.yx,menshuan.wyy,qingli.hql,zijiao.wzj\}@alibaba-inc.com} \\
  \texttt{\{wanglongyue.wly,weihua.luowh,kaifu.zkf\}@alibaba-inc.com} \\
  \texttt{\{hubaotian,zhangmin2021\}@hit.edu.cn} \\
  \faGithub~~\href{https://github.com/AIDC-AI/ComfyUI-Copilot}{\texttt{https://github.com/AIDC-AI/ComfyUI-Copilot}} \\
}

\begin{document}
\maketitle
\begin{abstract}
We introduce \textbf{ComfyUI-Copilot}, a large language model-powered plugin designed to enhance the usability and efficiency of ComfyUI, an open-source platform for AI-driven art creation. 
Despite its flexibility and user-friendly interface, ComfyUI can present challenges to newcomers, including limited documentation, model misconfigurations, and the complexity of workflow design.
ComfyUI-Copilot addresses these challenges by offering intelligent node and model recommendations, along with automated one-click workflow construction.  
At its core, the system employs a hierarchical multi-agent framework comprising a central assistant agent for task delegation and specialized worker agents for different usages,
supported by our curated ComfyUI knowledge bases to streamline debugging and deployment. 
We validate the effectiveness of ComfyUI-Copilot through both offline quantitative evaluations and online user feedback, 
showing that it accurately recommends nodes and accelerates workflow development. 
Additionally, use cases illustrate that ComfyUI-Copilot lowers entry barriers for beginners and enhances workflow efficiency for experienced users. 
The ComfyUI-Copilot installation package and a demo video are available at  
\url{https://github.com/AIDC-AI/ComfyUI-Copilot}.

\end{abstract}

\section{Introduction}
\input{1_intro}

\section{Related Work}

\begin{figure*}[!ht]
    \centering
    \includegraphics[width=\textwidth]{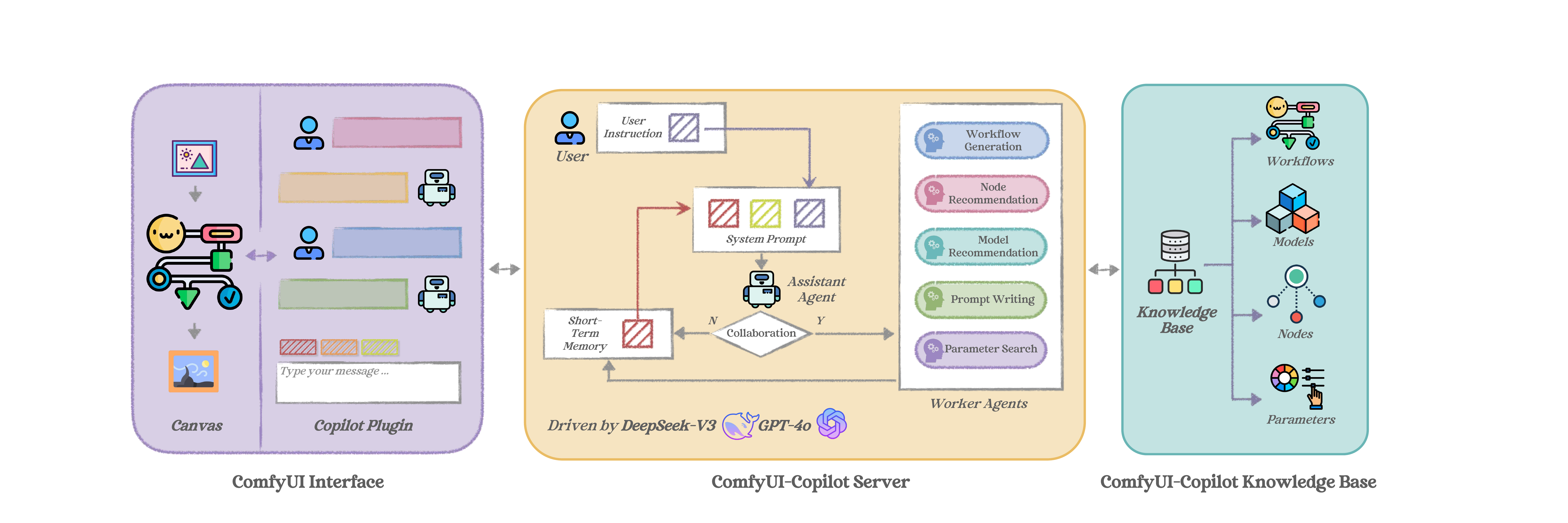}
    \caption{\textbf{Overview of the \ours framework}: The central LLM-based assistant agent can either respond directly to user instructions based on the conversation history (i.e., short-term memory), or collaborate with specialized worker agents. These agents are supported by our curated ComfyUI knowledge bases.}
    \label{fig:framework}
\end{figure*}

\input{2_related}

\section{ComfyUI-Copilot}
\input{3_method}

\section{Usage and Evaluation}

\input{4_exp}



\section{Conclusion}
In this paper, we present \ours, an LLM-powered multi-agent framework designed to
address ComfyUI-related queries and enable one-click workflow creation,
thereby lowering the barriers of ComfyUI development. 
By leveraging an LLM as a core assistant agent and integrating specialized worker agents and extensive knowledge bases, 
\ours not only enhances the workflow generation process with a high recall rate, but also ensures that it stays current with the latest modules in multimodal generation.
As the first project to explore a ComfyUI assistant plugin for providing instant suggestions,
\ours has rapidly gathered over 1.6K stars, attracted 19K users across 22 countries and processed more than 85K queries.
In future work, we plan to incorporate feedback from GitHub issues and actively update features to address user pain points, such as automatic workflow and parameter optimization.

\section*{Acknowledgments}

We want to thank anonymous reviewers for their helpful comments. 
This work is jointly
supported by grants: Natural Science Foundation of China (No. 62422603),
Guangdong Basic and Applied Basic Research Foundation (No. 2023A1515110078), and
Shenzhen Science and Technology Program (No. ZDSYS20230626091203008).

\bibliography{0_main}

\clearpage
\appendix

\input{5_appendix}


\end{document}

%% file: 1_intro.tex

Recent advancements in large language models (LLMs) and image generation methods have democratized AI-generated content (AIGC) production, 
with open-source frameworks like ComfyUI~\cite{comfyanonymous2023comfyui} emerging as pivotal tools for low-code AI workflow development. 
Serving over 4 million active users and backed by a vibrant community contributing 12K+ components (e.g., SDXL~\cite{podell2023sdxl}, ControlNet~\cite{zhang2023adding}), 
ComfyUI enables flexible workflow orchestration via drag-and-drop components for multimodal tasks such as text-to-image generation, face swapping, and video editing. 


\begin{figure*}[!ht]
    \centering
    \includegraphics[width=\textwidth]{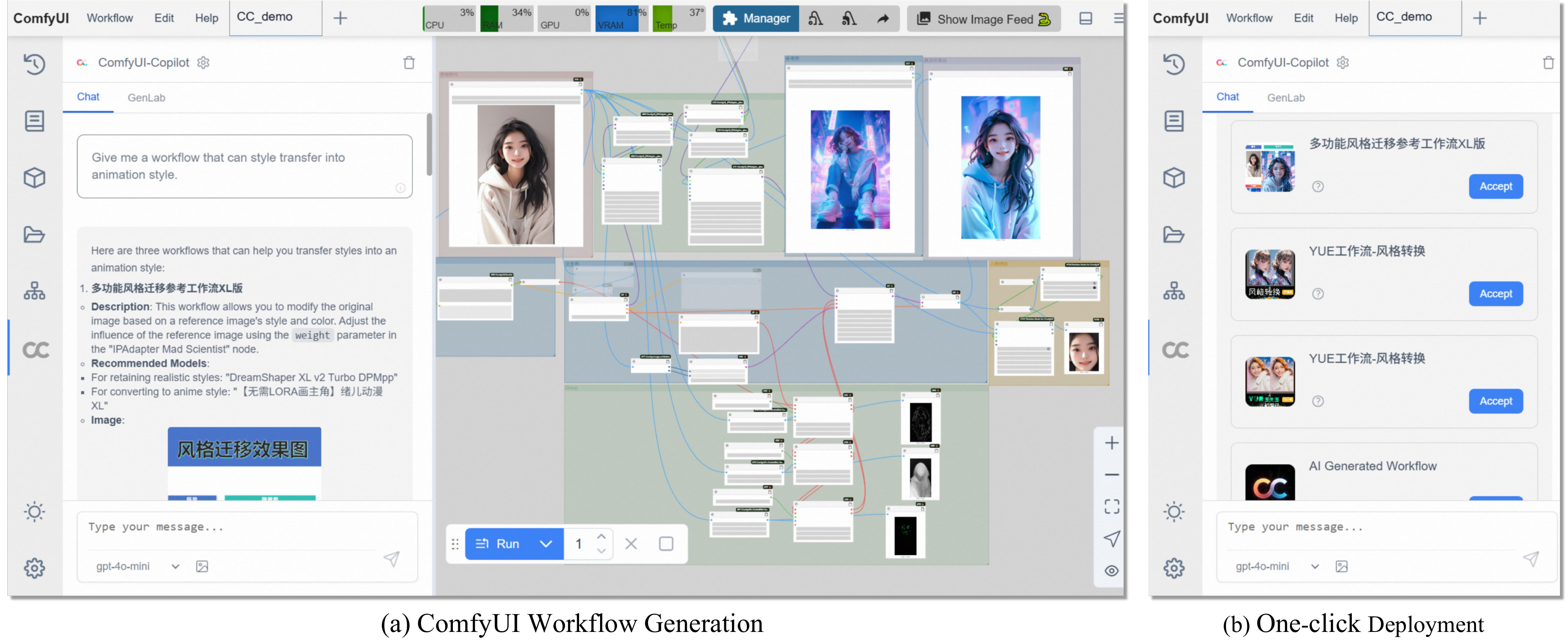}
    \caption{\textbf{An example of automated workflow generation in \ours}: the copilot suggests multiple workflows based on the user instruction and loads the selected one into the canvas with a single click.}
    \label{fig:intro}
\end{figure*}

Despite its convenience, newcomers may face several potential barriers when starting with ComfyUI.
These challenges include the installation of dependent nodes and models, scattered documentation across forums and GitHub issues.
Even experienced users require substantial expertise to debug and construct a well-designed workflow~\cite{gal2024comfygen}. 
Recent research on automatic workflow construction has limitations,
such as instability (i.e., generating unprocessable workflows) and a narrow focus primarily on text-to-image generation tasks~\cite{xue2024comfybench,sobania2024comfygi}.
Addressing these challenges and facilitating the onboarding process for ComfyUI is therefore crucial.

To this end,
we introduce \textbf{ComfyUI-Copilot},
an LLM-empowered multi-agent framework designed to assist users in navigating ComfyUI. 
It provides the following key features:
\textbf{(1) Automatic workflow generation:} 
Our copilot can identify user intent, retrieve or synthesize an appropriate workflow, and then integrate it into the ComfyUI canvas.
An example of its functionality is shown in Figure~\ref{fig:intro}.
\textbf{(2) Node and model recommendation:} 
Our copilot can suggest suitable nodes based on user instructions, recommend relevant checkpoints and LoRA models.
\textbf{(3) ComfyUI-related question answering:}
Our copilot provides detailed tutorials on selected nodes and models,
including usage guidelines, installation steps, and parameter explanations.
It can also offer multiple feasible downstream subgraphs for selected nodes.
Additionally, we introduce new features aimed at enhancing workflow debugging and optimization, 
including prompt writing and parameter search.

The framework of \ours is centered around an LLM-based assistant agent, which coordinates with various specialized worker agents and knowledge bases (KBs).
Depending on the query, the assistant agent may address user queries directly or delegate tasks to appropriate worker agents. 
We have developed three primary worker agents focused on workflow generation, node and model recommendation.
To support these agents, we have constructed extensive KBs covering 7K nodes, 62K models, and 9K workflows.
These KBs are enhanced through automated documentation generation by leveraging LLM's code comprehension capabilities, and are continuously expanded and updated daily.
Unlike prior work~\cite{gal2024comfygen,sobania2024comfygi} which only targets text-to-image generation, the resources in our KBs extend to conditional multimodal generation tasks, 
ensuring that our system accommodates both diverse tasks and cutting-edge modules with accuracy.

Experiments show that \ours provides accurate assistance in node recommendation and workflow construction based on user instructions.
The high recall rates for workflows and nodes (both exceeding 88.5\%) validate the practical efficacy in automated workflow development and accurate node recommendation.
Since its release on GitHub, online user feedback reflects a moderately high acceptance rate of 65.4\% for recommended nodes and a notably high acceptance rate of 85.9\% for proposed workflows. 
Use cases further highlight the system’s capability to reduce entry barriers for beginners and enhance workflow efficiency for experienced ComfyUI users with multilingual support.

To the best of our knowledge, 
\ours is the first open-source project to develop a ComfyUI plugin for automating workflow creation and providing instant suggestions. 
As of the camera-ready date (May 29, 2025), it has already attracted a rapidly growing user base, accumulating over 1.6K GitHub stars and processing more than 85K queries from 19K users across 22 countries. 
In future work, we plan to incorporate feedback from the active open-source community and continuously update features to better address user needs.

%% file: 2_related.tex
\textbf{AI-generated content (AIGC) based on ComfyUI.}
Diffusion models have gained wide attention in AI research for image synthesis~\cite{NEURIPS2020_4c5bcfec_diffusion,NEURIPS2021_49ad23d1_diffusion}. 
As the field of text-to-image generation progresses,
new tasks and models \citep{kumari2023multi,ruiz2023dreambooth, li2023dreamedit, zhang2023adding} have been proposed to introduce controllable conditions in image generation.
Therefore, researchers and practitioners are transitioning from simple text-to-image workflows to more sophisticated ones,
where the open-source ComfyUI~\cite{comfyanonymous2023comfyui} offers great convenience.
In ComfyUI, users can easily construct workflows by
connecting a series of blocks, each representing specific models or parameter choices.
Each ComfyUI workflow can be exported to a JSON file which outlines both the graph nodes and their connectivity.

Instead of relying on an end-to-end diffusion model for image generation, 
advanced workflows combine a variety of components to enhance image quality~\cite{guo2024pulid,ye2023ip}.
These components may include fine-tuned versions of generative models, large language models (LLMs) for refining input prompts, LoRAs trained to introduce specific artistic styles, improved latent decoders for finer details, super-resolution blocks, and more~\citep{hu2021lora,manas2024texttoimage,berrada2025boosting,super_resolution}.
Importantly, effective workflows are prompt-dependent,
with the selection of models and nodes often based on the user intent and the desired image content~\cite{gal2024comfygen}.
Therefore, creating a well-designed workflow and selecting appropriate nodes and models require significant expertise,
where our \ours comes into help.

\noindent \textbf{LLM-based agents.}
Recent advancements in LLMs have demonstrated great improvements in reasoning abilities and adaptability to new content and tasks~\cite{chen-etal-2024-agent,zeng-etal-2024-agenttuning,wang2025cigeval}. 
Based on these emergent capabilities~\cite{wei2022emergent},
various studies have utilized LLMs for agentic task completion using external tools, including hallucination detection~\cite{cheng-etal-2024-small}, visual question answering~\cite{cheng-etal-2024-least,yin-etal-2024-agent}, and web navigation~\cite{agashe2025agents,yang2025agentoccam,li2025perceptionreasonthinkplan}.
In addition to tools, LLM-based agents are often equipped with components 
such as memory mechanisms~\cite{wang2024agentworkflowmemory,xu-etal-2025-mekb}, retrieval modules~\cite{asai2024selfrag,kim-etal-2024-rada} and reasoning strategies like self-reflection~\cite{shinn2023reflexion,xu2023reasoning},
aimed at enhancing their overall performance.

Our work proposes a multi-agent framework for 
the automated development and deployment of ComfyUI workflows.
In this framework,
the LLM acts as the central planner, 
autonomously selecting suitable worker agents to address diverse user queries. 
Although recent research has shown increasing interest in workflow generation~\cite{gal2024comfygen,xue2024comfybench,sobania2024comfygi},
existing methods often face challenges such as instability, leading to unparseable output workflows, or are limited to text-to-image tasks. 
We broaden the scope to include various conditional image and video generation tasks,
and address user queries with a high acceptance rate.

%% file: 3_method.tex
In this section, we provide a detailed description of \ours.
As illustrated in Figure~\ref{fig:framework},
the system utilizes a hierarchical multi-agent framework that
includes a central assistant agent for task delegation and specialized worker agents for different usages.
We first introduce our curated ComfyUI knowledge bases (Sec.~\ref{sec:kb}),
and the details of the multi-agent framework (Sec.~\ref{sec:agent}).
Following this, we present the interactive chat interface and provide usage examples in Section~\ref{sec:interface}.

\subsection{Knowledge Bases}
\label{sec:kb}

We have constructed three KBs about nodes, models and workflows.
The data is sourced from popular platforms for sharing generative resources, ComfyUI-related GitHub repositories, and the ComfyUI website, with NSFW content filtered out.

\begin{figure*}[!ht]
    \centering
    \includegraphics[width=\textwidth]{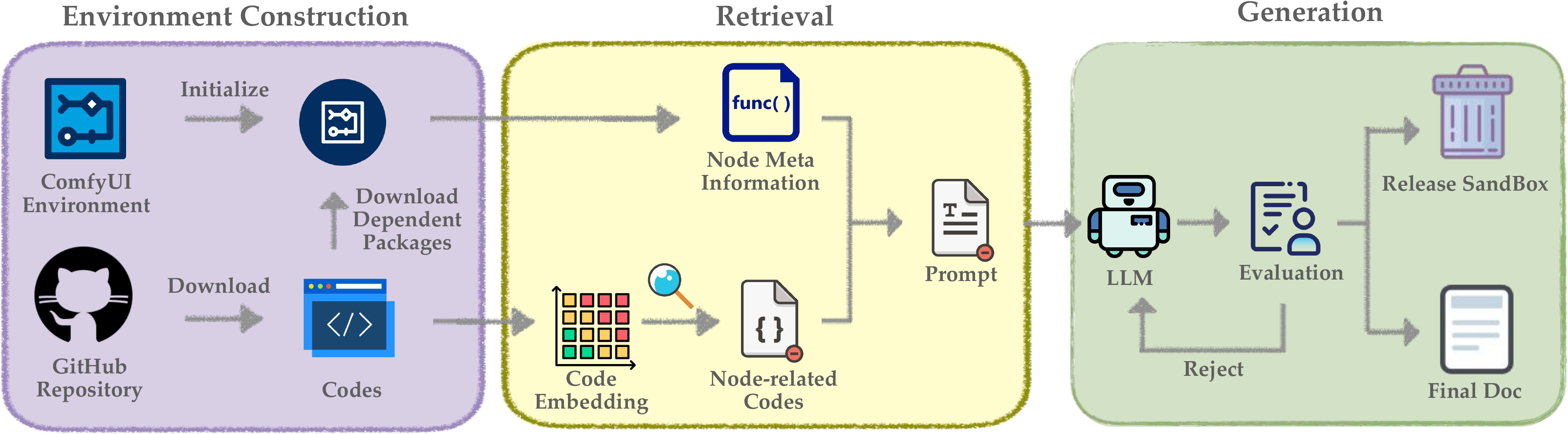}
    \caption{\textbf{The process of automatic node documentation generation.}  Starting from GitHub repositories, the process involves constructing an executable ComfyUI environment, followed by code chunking and retrieval, and concludes with generating the final documentation.}
    \label{fig:code}
\end{figure*}

For nodes lacking structured documentation,
we automatically generate detailed documentation by analyzing their GitHub repositories.
As shown in Figure~\ref{fig:code}, 
the process begins by setting up a sandbox environment to run ComfyUI, cloning the GitHub repositories, and installing the necessary dependencies. 
After successfully importing the nodes within ComfyUI, 
we extract metadata, including the node class type, input and output parameters.
The GitHub code is then segmented into chunks, which are embedded using the BGE-M3 embedding~\cite{chen2024bgem3embeddingmultilingualmultifunctionality}, followed by retrieval to locate relevant code for each node.
By combining the metadata with the code,
we use an LLM to 
generate documentation on node usage and parameter meanings.
The generated documentation undergoes quality reviews before finalization, with an example provided in Appendix~\ref{sec:app_node}.

In addition,
since community-sourced content tends to focus more on installation instructions, there is often a lack of detailed explanations of workflow and model functionalities.
To address this, we leverage the multimodal understanding capabilities of GPT-4o,
by prompting it with community-sourced texts, accompanying images that typically demonstrate the effects of the workflows or models, and any available workflow JSON files.
This approach helps fill in the gaps in usage descriptions, which is essential for further developing effective recommendation worker agents.

In total, we have constructed extensive KBs covering 7K nodes, 62K models, and 9K workflows.
These KBs are continuously expanded weekly, 
covering a wide range of conditional image and video generation tasks.
This ensures that \ours remains adaptable to both widely used and cutting-edge modules.

\subsection{Agents}
\label{sec:agent}

The core of \ours is a well-instructed LLM-based assistant agent, serving as a planner.
Depending on the user instruction,
the assistant either responds to queries using the constructed KBs or delegates tasks to appropriate worker agents. 
We have created three worker agents for workflow, nodes and models,
which we collectively refer to as ``modules'' in this section.
The recommendation process for each module follows a three-stage pipeline, 
progressing from coarse to fine granularity. 

In the first stage, we employ an LLM or a large multimodal model (LMM), such as DeepSeek-V3 or GPT-4o, to expand vague user instructions into detailed task descriptions and noteworthy considerations.
For example, when performing style transfer, if the LMM identifies the original image as a human portrait, the expanded user intent will highlight the importance of maintaining subject consistency.
In the second stage, we represent the user intent as an embedding and calculate its cosine distance with modules in the KB, obtaining a semantic score $\text{sim}_S$ based on OpenAI's \texttt{text-embedding-3-small}.
Additionally, we compute a lexical similarity score $\text{sim}_L$ based on the proportion of overlapping words.
The overall retrieval score $\text{sim}_O$ is calculated as:
\begin{equation}
\text{sim}_O = 0.7 \times \text{sim}_S + 0.3 \times \text{sim}_L
\end{equation}
The top 30 modules with the highest $\text{sim}_O$ scores are then selected for further re-ranking.
In the third stage, we use the GTE-Rerank model\footnote{\url{https://huggingface.co/Alibaba-NLP/gte-multilingual-reranker-base}} to determine the top 3 modules from the above candidates.
The re-ranking score is obtained by providing the re-ranker with the user intent and the description of each candidate module.
These top 3 modules are further ranked by considering popularity factors such as upvotes, downloads, and star statistics.

\begin{figure}[t]
    \centering
    \includegraphics[width=\linewidth]{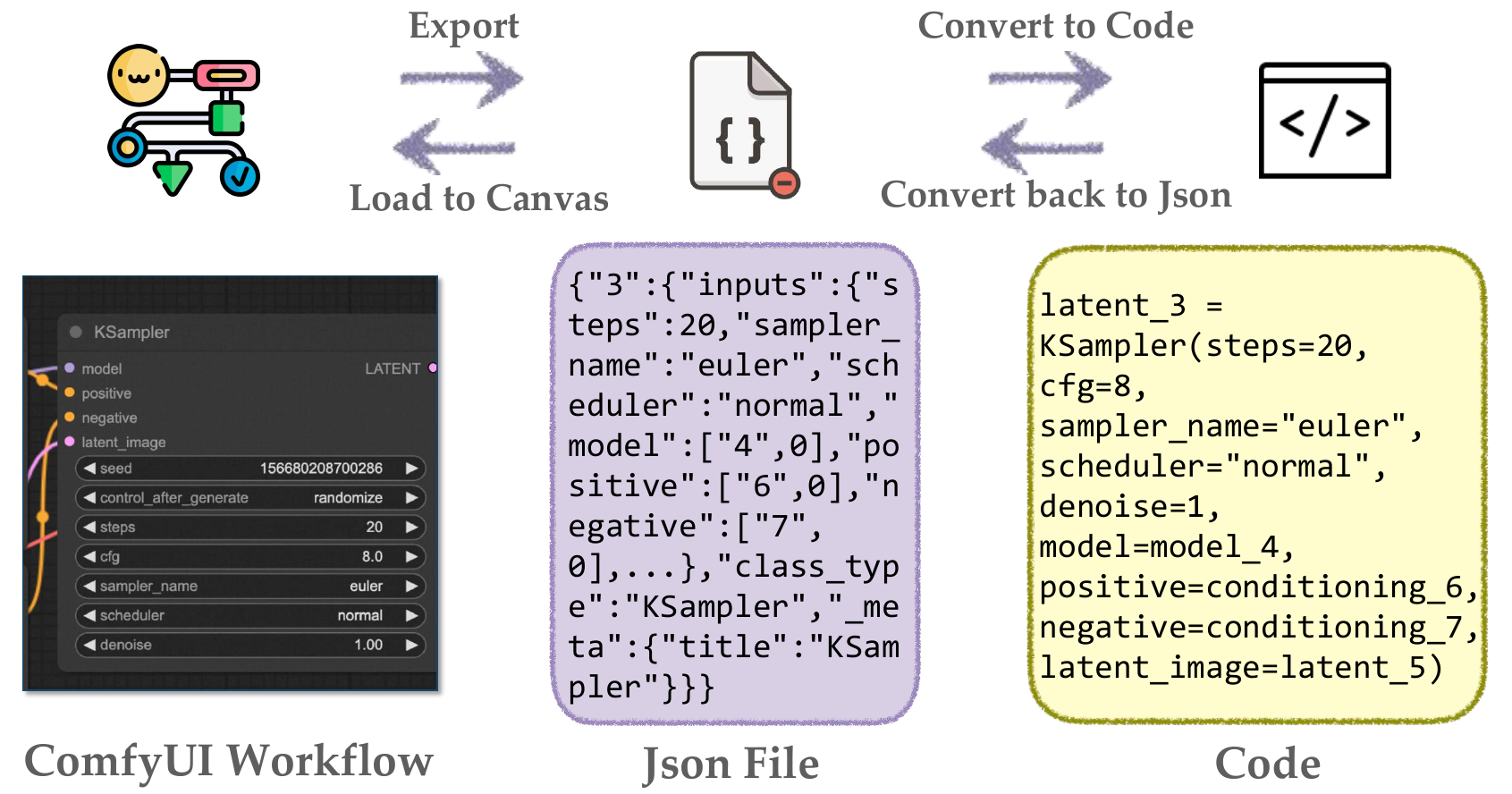}
    \caption{Different representations of ComfyUI workflows and their flexible conversions.}
    \label{fig:workflow}
\end{figure}

For the workflow generation agent,
in addition to the module recalling pipeline, 
we explore the possibility of generating workflows from scratch based on code LLMs.
As illustrated in Figure~\ref{fig:workflow},
workflows can be represented in three common formats: ComfyUI flow graphs, JSON, and code.
Following~\citet{xue2024comfybench}, 
we enable mutual conversion between the JSON and code formats based on graph topology using Python-like syntax.
We adopt code as the primary workflow representation due to its rich logical and semantic information, as well as its natural compatibility with LLMs' code generation capabilities.
Given a user instruction, we prompt top-tier closed-source LLMs with retrieved nodes and code exemplars to generate workflows from scratch.
Additionally, to investigate whether task-specific open-source models can replace closed-source LLMs, 
we fine-tune open-sourced Qwen2.5-Coder-7B~\cite{hui2024qwen25codertechnicalreport} on workflows collected in our KB.
Experimental results in Table~\ref{tab:workflow} show that the fine-tuned model achieves performance comparable to Claude-3.7-Sonnet in terms of pass rate and node selection in generated workflows. 
More evaluation details are in Appendix~\ref{sec:app_workflow}.
However, due to the inherent complexity of workflow generation~\cite{gal2024comfygen}, 
there remains significant room for improvement in pass rates.

Implemented with LangChain\footnote{\url{https://www.langchain.com/}},
our framework (Figure~\ref{fig:framework}) equips the assistant agent to autonomously select appropriate worker agents based on user instructions and short-term memory (i.e., message history). 
The assistant then synthesizes responses by integrating outputs from these worker agents, 
enabling automated ComfyUI-related question answering, workflow generation, and module recommendation. 
For prompt writing and parameter search functionality, we provide illustrative examples in Section~\ref{sec:interface}.


\subsection{Interface}
\label{sec:interface}

As shown in Figure~\ref{fig:intro}, \ours is seamlessly integrated into the ComfyUI interface. 
Users can launch our service with a single click on the \ours icon in the left sidebar. 
Once activated, the chat box displays user inputs and our copilot's responses.
Users can engage in multiple rounds of conversation and switch between underlying LLMs such as DeepSeek-V3 and GPT-4o.

\begin{figure*}[!ht]
    \centering
    \includegraphics[width=0.9\textwidth]{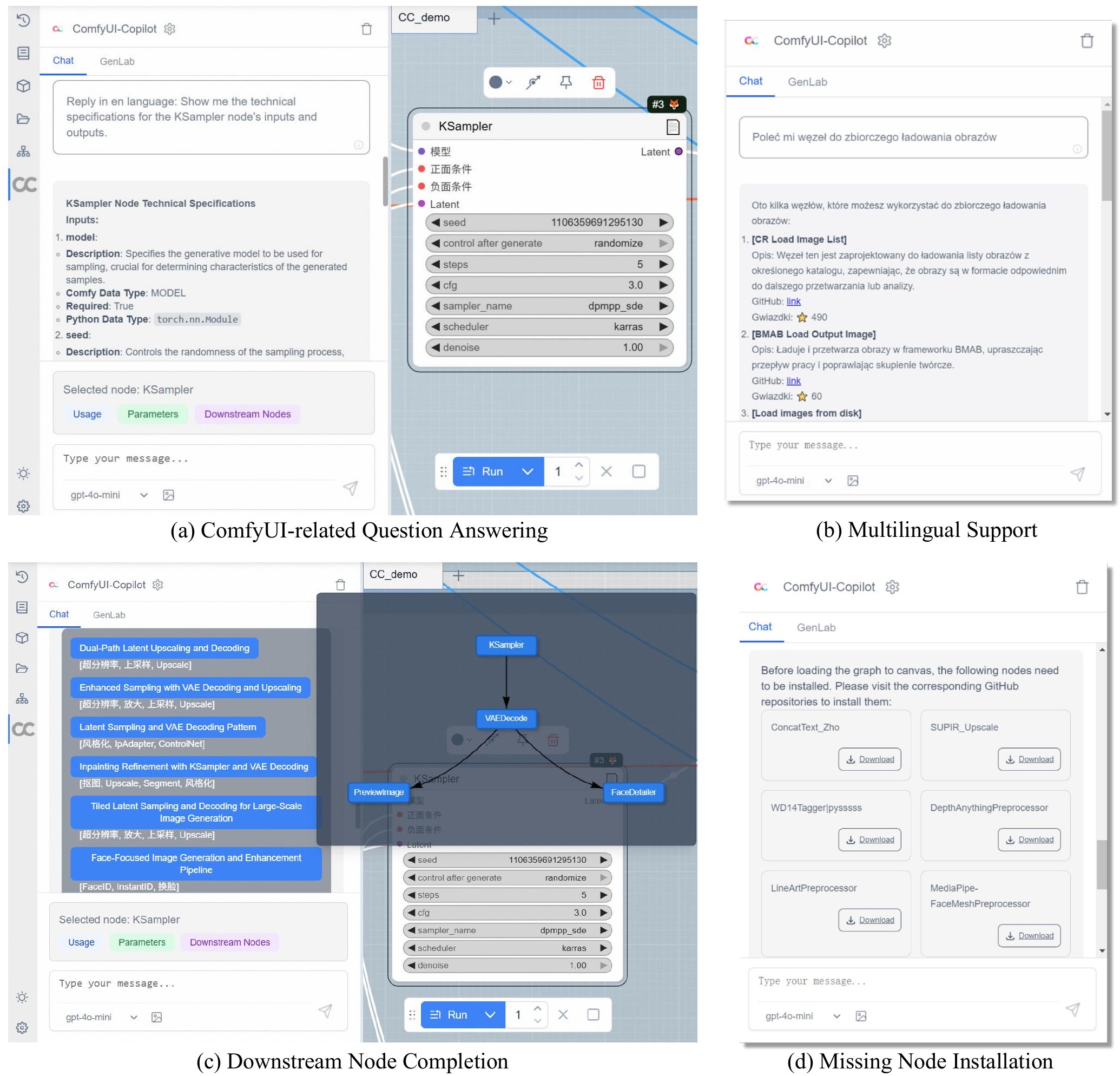}
    \caption{Examples of \ours's different usages.}
    \label{fig:main_case}
\end{figure*}

\textbf{Automatic Workflow Generation.}
As illustrated in Figure~\ref{fig:intro},
\ours responds to user instructions by presenting the top three recalled workflows.
By clicking ``Accept'',
the selected workflow can be loaded onto the canvas.
If \ours detects that any required nodes are missing,
it provides an installation guide and directs the user to the official GitHub repositories for easy setup (See Figure~\ref{fig:main_case} (d)).

\textbf{ComfyUI-related Question Answering.} 
Users can click on any node to ask shortcut questions about its usage, parameters, and recommended downstream nodes. 
Figure~\ref{fig:main_case} (a) and (c) illustrate this feature: a user inquires about the input and output parameters of the ``KSampler'' node, and \ours not only explains them but also suggests relevant downstream nodes, such as subgraphs for face swapping and image upscaling, to streamline workflow construction.
Additionally, as shown in Figure~\ref{fig:main_case} (b), 
\ours supports multilingual queries and responses (e.g., Polish in the example), enhancing accessibility for users worldwide.

\textbf{Node and Model Recommendation.}
Module recommendations in \ours are context-aware, 
taking into account dependencies between components in the workflow. 
For instance, certain LoRA models perform optimally with specific diffusion models. 
As shown in Figure~\ref{fig:app_case} (a), when a user requests a LoRA model for text-to-image generation, 
\ours prompts the user to specify the diffusion model being used before suggesting compatible LoRA models.
Figure~\ref{fig:app_case} (b) demonstrates an example of node recommendation. 
The interface displays detailed descriptions and GitHub star counts for each recommended node, 
allowing users to add their preferred choice to the canvas with a single click.


\begin{table}[t]
\centering
\begin{tabular}{lcc}
\toprule
  & \textbf{DeepSeek-V3} & \textbf{GPT-4o}  \\
\midrule
Node  & 0.885 & 0.894 \\
Workflow  & 0.900 & 0.892  \\
\bottomrule
\end{tabular}
\caption{Recall rates of nodes and workflows in \ours based on our constructed test set.}
\label{tab:offline_test}
\end{table}

In addition to the core features that lower entry barrier for beginners, 
we also provide new features to enhance productivity for experienced ComfyUI users.
The prompt-writing functionality in Figure~\ref{fig:prompt_writing} helps users refine prompts for text-to-image generation, resulting in more vivid images.
For example, given a simple instruction like ``a cat'', several detailed prompts are proposed, each leading to high-quality outputs.
Figure~\ref{fig:parameter} illustrates the parameter search functionality, which enables users to run parallel experiments by varying key parameters and batch-processing images for efficient comparison. 
In the given example, the image generated using the original workflow does not resemble the source sofa image. 
By experimenting with different combinations of parameters (specifically ``cfg'' and ``denoise'' in the KSampler node),
the resulting images can be compared side by side, allowing users to easily identify the optimal parameters that best preserve the desired attributes.


%% file: 4_exp.tex
To evaluate the performance of \ours, 
we designed 130 user instructions for workflow recall based on our workflow KB.
These instructions are created by rewriting the usage descriptions of specific workflows, using the target workflow as the correct answer,
Examples include ``I need a workflow that is suitable for fast upscaling and image quality restoration''.
Similarly, We create 104 node recommendation instructions based on our node KB, such as ``I want to enhance image aesthetics and resolution in AI art applications, recommend a suitable node''.

As shown in Table~\ref{tab:offline_test}, 
when recalling the top three workflows and nodes, our framework achieves high recall rates (over 88.5\%) with both GPT-4o and DeepSeek-V3.
This demonstrates the robustness and effectiveness of our multi-agent framework. 
Error analysis on the unsuccessful workflow cases indicates that, 
even when the exact target workflow is not recalled, 
the suggested workflows often still fulfill the user's intended functions.

Since releasing \ours on GitHub on February 23, 2025,  
online user feedback has shown a moderately high acceptance rate of
65.4\% for recommended nodes 
and a notably high acceptance rate of 85.9\% for proposed workflows.
As the first open-source project for a ComfyUI assistant plugin,
\ours has quickly attracted a growing user base with active engagement,
received over 1.6K Github stars, with 85K queries from 19K users across 22 countries.
Thanks to the open-source community, we have gathered valuable feedback from GitHub issues and are actively updating features to better address user needs.

%% file: 5_appendix.tex
\section{Example of Node Documentation}
\label{sec:app_node}

Here we present an example of automatic node documentation generation using GPT-4o.
The input GitHub repository is ComfyUI-SUPIR~\cite{yu2024scaling}\footnote{\url{https://github.com/kijai/ComfyUI-SUPIR}}.
The resulting documentation is as follows.

\NewTColorBox{ToolUse_Box}{ s O{!htbp} }{%
  floatplacement={#2},
  IfBooleanTF={#1}{float*,width=\textwidth}{float},
  colframe=cyan!50!black,colback=cyan!10!white,title=SUPIR Upscale Documentation,
  }

\begin{ToolUse_Box}[!h]

The SUPIR\_Upscale node is designed to enhance image resolution using advanced upscaling techniques, leveraging the SUPIR and SDXL models for high-quality output. It allows for various configurations, including different upscaling methods and model parameters, to optimize the image enhancement process.

\#\# Input types

\begin{itemize}
    \item \textbf{\texttt{supir\_model}}
        \begin{itemize}
            \item Specifies the path to the SUPIR model, which is essential for the upscaling process, ensuring that the node can utilize the trained model for image enhancement.
            \item Type: \texttt{COMBO[STRING]}
        \end{itemize}
    \item \textbf{\texttt{sdxl\_model}}
        \begin{itemize}
            \item Indicates the path to the SDXL model, which works in conjunction with the SUPIR model to improve the quality of the upscaled images.
            \item Type: \texttt{COMBO[STRING]}
        \end{itemize}
    \item  \textbf{(More inputs omitted)}
\end{itemize}

\#\# Output types
\begin{itemize}
    \item \textbf{\texttt{upscaled\_image}}
        \begin{itemize}
            \item The resulting image after the upscaling process, enhanced in resolution and quality based on the input parameters.
            \item Type: \texttt{IMAGE}
        \end{itemize}
\end{itemize}
\end{ToolUse_Box}

\section{Automatic Workflow Generation Experiment}
\label{sec:app_workflow}

In this experiment,
we randomly select 2K high-quality workflows from the KB for training and 100 for evaluation.
The training data's input includes workflow usage, retrieved nodes, and code examples.
The code representation of the target workflow is the desired output.
We fine-tune Qwen2.5-Coder-7B~\cite{hui2024qwen25codertechnicalreport} with LLaMA-Factory~\cite{zheng-etal-2024-llamafactory}.
The fine-tuning process employs a learning rate of 1e-5 and a batch size of 16, with a sequence length of 16K.

We compare the fine-tuned Qwen2.5-Coder-7B with the retrieval-augmented method based on closed-source models such as GPT-4o and Claude-3.7-Sonnet.
Evaluation metrics include the pass rate (i.e., whether the generated workflow can be executed within the ComfyUI canvas), the average number of nodes, and the precision, recall, and F1 score for node selection.
Results in Table~\ref{tab:workflow} show that our fine-tuned model performs comparably to Claude-3.7-Sonnet, achieving the highest F1 score for node selection (0.95). 
Although GPT-4o achieves the highest pass rate, a closer examination reveals that it tends to generate overly simplistic workflows (an average of 8 nodes). 83\% of the workflows produced by GPT-4o contain fewer nodes than the target workflows, leading to low node recall rates.
Despite the promising performance of our fine-tuned model and Claude-3.7-Sonnet, there remains significant room for further improvements in workflow generation.

\begin{table}[h]
\small
\centering
\begin{tabular}{lccccc}
\toprule
\multirow{2}{*}{\textbf{Model}} & \multirow{2}{*}{\textbf{Pass}} & \multirow{2}{*}{\textbf{\#Nodes}} & \multicolumn{3}{c}{\textbf{Node}} \\
\cmidrule(lr){4-6}
& & & \textbf{P} & \textbf{R} & \textbf{F1} \\
\midrule
GPT-4o & \textbf{0.92} & 8 & 0.91 & 0.65 & 0.75 \\
Claude 3.7 &  {0.73} &  {13} &  {0.90} &  {0.88} &  {0.88} \\
Ours & 0.74 & \textbf{14} & \textbf{0.96} & \textbf{0.94} & \textbf{0.95} \\
\bottomrule
\end{tabular}
\caption{Performance comparison of LLMs for workflow generation across evaluation metrics. \#Nodes means the number of nodes. Precision (P), recall (R), and F1-score at node level are reported.}
\label{tab:workflow}
\end{table}


\begin{figure*}[!ht]
    \centering
    \includegraphics[width=\textwidth]{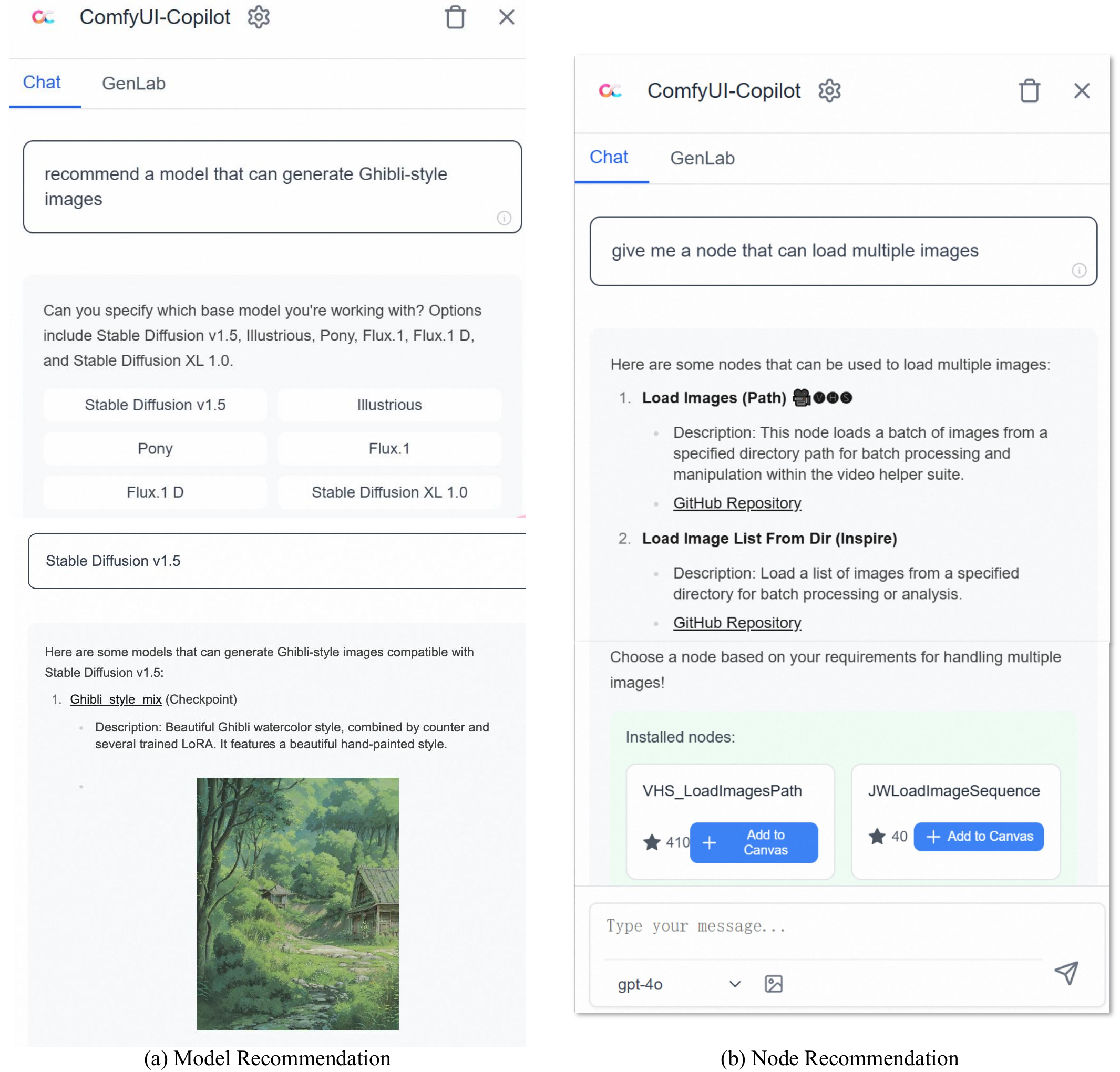}
    \caption{Model and node recommendation in \ours.}
    \label{fig:app_case}
\end{figure*}

\begin{figure*}[!ht]
    \centering
    \includegraphics[width=\textwidth]{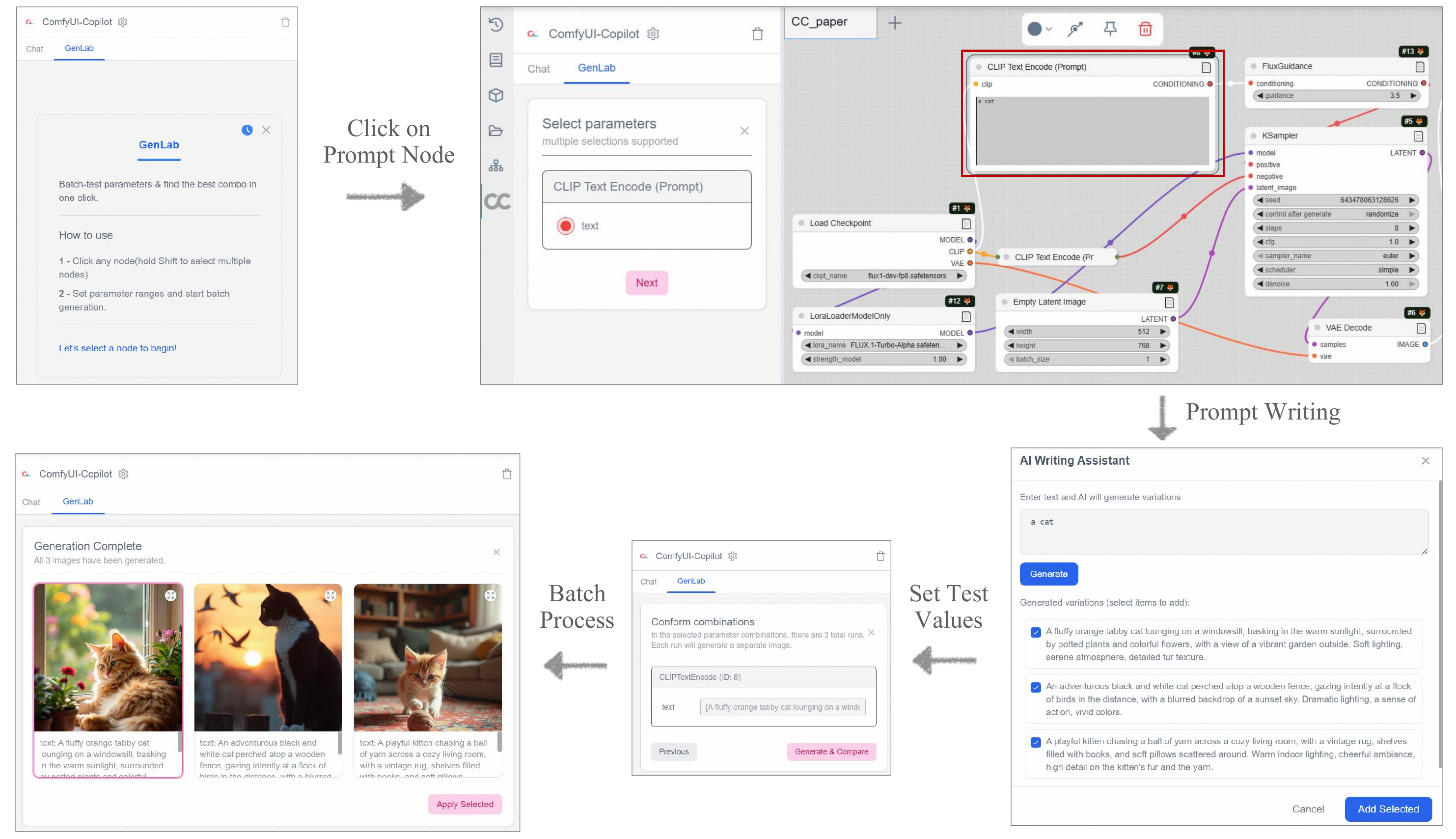}
    \caption{Prompt writing in \ours.}
    \label{fig:prompt_writing}
\end{figure*}

\begin{figure*}[!ht]
    \centering
    \includegraphics[width=0.95\textwidth]{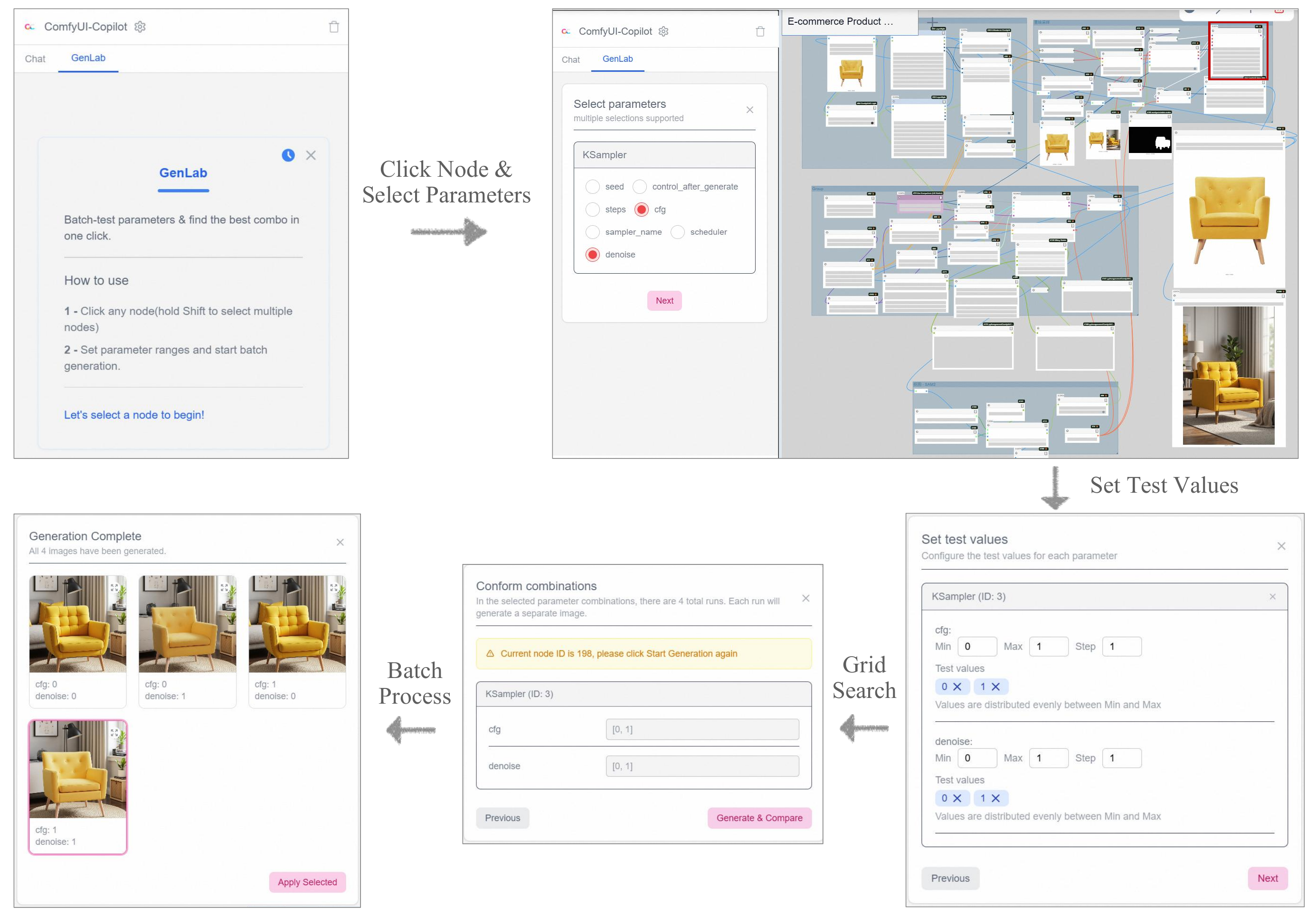}
    \caption{Parameter search in \ours.}
    \label{fig:parameter}
\end{figure*}

\section{More Examples of \ours}

Due to page limit, we demonstrate the remaining functionalities in Figure~\ref{fig:app_case},~\ref{fig:prompt_writing} and \ref{fig:parameter}, 
including node and model recommendation, prompt writing assistance, and parameter search.

%% file: 0_main.bbl
\begin{thebibliography}{37}
\providecommand{\natexlab}[1]{#1}

\bibitem[{Agashe et~al.(2025)Agashe, Han, Gan, Yang, Li, and Wang}]{agashe2025agents}
Saaket Agashe, Jiuzhou Han, Shuyu Gan, Jiachen Yang, Ang Li, and Xin~Eric Wang. 2025.
\newblock \href {https://openreview.net/forum?id=lIVRgt4nLv} {Agent s: An open agentic framework that uses computers like a human}.
\newblock In \emph{The Thirteenth International Conference on Learning Representations}.

\bibitem[{Asai et~al.(2024)Asai, Wu, Wang, Sil, and Hajishirzi}]{asai2024selfrag}
Akari Asai, Zeqiu Wu, Yizhong Wang, Avirup Sil, and Hannaneh Hajishirzi. 2024.
\newblock \href {https://openreview.net/forum?id=hSyW5go0v8} {Self-{RAG}: Learning to retrieve, generate, and critique through self-reflection}.
\newblock In \emph{The Twelfth International Conference on Learning Representations}.

\bibitem[{Berrada et~al.(2025)Berrada, Astolfi, Hall, Havasi, Benchetrit, Romero-Soriano, Alahari, Drozdzal, and Verbeek}]{berrada2025boosting}
Tariq Berrada, Pietro Astolfi, Melissa Hall, Marton Havasi, Yohann Benchetrit, Adriana Romero-Soriano, Karteek Alahari, Michal Drozdzal, and Jakob Verbeek. 2025.
\newblock \href {https://openreview.net/forum?id=y4DtzADzd1} {Boosting latent diffusion with perceptual objectives}.
\newblock In \emph{The Thirteenth International Conference on Learning Representations}.

\bibitem[{Chen et~al.(2024{\natexlab{a}})Chen, Xiao, Zhang, Luo, Lian, and Liu}]{chen2024bgem3embeddingmultilingualmultifunctionality}
Jianlv Chen, Shitao Xiao, Peitian Zhang, Kun Luo, Defu Lian, and Zheng Liu. 2024{\natexlab{a}}.
\newblock \href {https://arxiv.org/abs/2402.03216} {Bge m3-embedding: Multi-lingual, multi-functionality, multi-granularity text embeddings through self-knowledge distillation}.
\newblock \emph{Preprint}, arXiv:2402.03216.

\bibitem[{Chen et~al.(2024{\natexlab{b}})Chen, Liu, Wang, Zhang, Liu, Lin, Chen, and Zhao}]{chen-etal-2024-agent}
Zehui Chen, Kuikun Liu, Qiuchen Wang, Wenwei Zhang, Jiangning Liu, Dahua Lin, Kai Chen, and Feng Zhao. 2024{\natexlab{b}}.
\newblock \href {https://doi.org/10.18653/v1/2024.findings-acl.557} {Agent-{FLAN}: Designing data and methods of effective agent tuning for large language models}.
\newblock In \emph{Findings of the Association for Computational Linguistics: ACL 2024}, pages 9354--9366, Bangkok, Thailand. Association for Computational Linguistics.

\bibitem[{Cheng et~al.(2024{\natexlab{a}})Cheng, Guan, Wu, and Yan}]{cheng-etal-2024-least}
Chuanqi Cheng, Jian Guan, Wei Wu, and Rui Yan. 2024{\natexlab{a}}.
\newblock \href {https://doi.org/10.18653/v1/2024.emnlp-main.284} {From the least to the most: Building a plug-and-play visual reasoner via data synthesis}.
\newblock In \emph{Proceedings of the 2024 Conference on Empirical Methods in Natural Language Processing}, pages 4941--4957, Miami, Florida, USA. Association for Computational Linguistics.

\bibitem[{Cheng et~al.(2024{\natexlab{b}})Cheng, Li, Zhao, Zhang, Zhang, Zhang, Gai, and Wen}]{cheng-etal-2024-small}
Xiaoxue Cheng, Junyi Li, Xin Zhao, Hongzhi Zhang, Fuzheng Zhang, Di~Zhang, Kun Gai, and Ji-Rong Wen. 2024{\natexlab{b}}.
\newblock \href {https://doi.org/10.18653/v1/2024.emnlp-main.809} {Small agent can also rock! empowering small language models as hallucination detector}.
\newblock In \emph{Proceedings of the 2024 Conference on Empirical Methods in Natural Language Processing}, pages 14600--14615, Miami, Florida, USA. Association for Computational Linguistics.

\bibitem[{comfyanonymous(2023)}]{comfyanonymous2023comfyui}
comfyanonymous. 2023.
\newblock Comfyui.
\newblock \url{https://github.com/comfyanonymous/ComfyUI}.

\bibitem[{Dhariwal and Nichol(2021)}]{NEURIPS2021_49ad23d1_diffusion}
Prafulla Dhariwal and Alexander Nichol. 2021.
\newblock \href {https://proceedings.neurips.cc/paper_files/paper/2021/file/49ad23d1ec9fa4bd8d77d02681df5cfa-Paper.pdf} {Diffusion models beat gans on image synthesis}.
\newblock In \emph{Advances in Neural Information Processing Systems}, volume~34, pages 8780--8794. Curran Associates, Inc.

\bibitem[{Gal et~al.(2024)Gal, Haviv, Alaluf, Bermano, Cohen-Or, and Chechik}]{gal2024comfygen}
Rinon Gal, Adi Haviv, Yuval Alaluf, Amit~H. Bermano, Daniel Cohen-Or, and Gal Chechik. 2024.
\newblock \href {https://arxiv.org/abs/2410.01731} {Comfygen: Prompt-adaptive workflows for text-to-image generation}.
\newblock \emph{Preprint}, arXiv:2410.01731.

\bibitem[{Guo et~al.(2024)Guo, Wu, Chen, chen, Zhang, and HE}]{guo2024pulid}
Zinan Guo, Yanze Wu, Zhuowei Chen, Lang chen, Peng Zhang, and Qian HE. 2024.
\newblock \href {https://openreview.net/forum?id=E6ZodZu0HQ} {Pu{LID}: Pure and lightning {ID} customization via contrastive alignment}.
\newblock In \emph{The Thirty-eighth Annual Conference on Neural Information Processing Systems}.

\bibitem[{Ho et~al.(2020)Ho, Jain, and Abbeel}]{NEURIPS2020_4c5bcfec_diffusion}
Jonathan Ho, Ajay Jain, and Pieter Abbeel. 2020.
\newblock \href {https://proceedings.neurips.cc/paper_files/paper/2020/file/4c5bcfec8584af0d967f1ab10179ca4b-Paper.pdf} {Denoising diffusion probabilistic models}.
\newblock In \emph{Advances in Neural Information Processing Systems}, volume~33, pages 6840--6851. Curran Associates, Inc.

\bibitem[{Hu et~al.(2021)Hu, Wallis, Allen-Zhu, Li, Wang, Wang, Chen et~al.}]{hu2021lora}
Edward~J Hu, Phillip Wallis, Zeyuan Allen-Zhu, Yuanzhi Li, Shean Wang, Lu~Wang, Weizhu Chen, et~al. 2021.
\newblock Lora: Low-rank adaptation of large language models.
\newblock In \emph{International Conference on Learning Representations}.

\bibitem[{Hui et~al.(2024)Hui, Yang, Cui, Yang, Liu, Zhang, Liu, Zhang, Yu, Lu, Dang, Fan, Zhang, Yang, Men, Huang, Zheng, Miao, Quan, Feng, Ren, Ren, Zhou, and Lin}]{hui2024qwen25codertechnicalreport}
Binyuan Hui, Jian Yang, Zeyu Cui, Jiaxi Yang, Dayiheng Liu, Lei Zhang, Tianyu Liu, Jiajun Zhang, Bowen Yu, Keming Lu, Kai Dang, Yang Fan, Yichang Zhang, An~Yang, Rui Men, Fei Huang, Bo~Zheng, Yibo Miao, Shanghaoran Quan, Yunlong Feng, Xingzhang Ren, Xuancheng Ren, Jingren Zhou, and Junyang Lin. 2024.
\newblock \href {https://arxiv.org/abs/2409.12186} {Qwen2.5-coder technical report}.
\newblock \emph{Preprint}, arXiv:2409.12186.

\bibitem[{Kim et~al.(2024)Kim, Bursztyn, Koh, Guo, and Hwang}]{kim-etal-2024-rada}
Minsoo Kim, Victor Bursztyn, Eunyee Koh, Shunan Guo, and Seung-won Hwang. 2024.
\newblock \href {https://doi.org/10.18653/v1/2024.findings-acl.802} {{R}a{DA}: Retrieval-augmented web agent planning with {LLM}s}.
\newblock In \emph{Findings of the Association for Computational Linguistics: ACL 2024}, pages 13511--13525, Bangkok, Thailand. Association for Computational Linguistics.

\bibitem[{Kumari et~al.(2023)Kumari, Zhang, Zhang, Shechtman, and Zhu}]{kumari2023multi}
Nupur Kumari, Bingliang Zhang, Richard Zhang, Eli Shechtman, and Jun-Yan Zhu. 2023.
\newblock Multi-concept customization of text-to-image diffusion.
\newblock In \emph{Proceedings of the IEEE/CVF Conference on Computer Vision and Pattern Recognition}, pages 1931--1941.

\bibitem[{Li et~al.(2023)Li, Ku, Wei, and Chen}]{li2023dreamedit}
Tianle Li, Max Ku, Cong Wei, and Wenhu Chen. 2023.
\newblock Dreamedit: Subject-driven image editing.
\newblock \emph{arXiv preprint arXiv:2306.12624}.

\bibitem[{Li et~al.(2025)Li, Liu, Li, Zhang, Xu, Chen, Shi, Jiang, Wang, Wang, Huang, Zhao, Jiang, Hong, Wang, Tian, Huai, Luo, Luo, Zhang, Hu, and Zhang}]{li2025perceptionreasonthinkplan}
Yunxin Li, Zhenyu Liu, Zitao Li, Xuanyu Zhang, Zhenran Xu, Xinyu Chen, Haoyuan Shi, Shenyuan Jiang, Xintong Wang, Jifang Wang, Shouzheng Huang, Xinping Zhao, Borui Jiang, Lanqing Hong, Longyue Wang, Zhuotao Tian, Baoxing Huai, Wenhan Luo, Weihua Luo, Zheng Zhang, Baotian Hu, and Min Zhang. 2025.
\newblock \href {https://arxiv.org/abs/2505.04921} {Perception, reason, think, and plan: A survey on large multimodal reasoning models}.
\newblock \emph{Preprint}, arXiv:2505.04921.

\bibitem[{Mañas et~al.(2024)Mañas, Astolfi, Hall, Ross, Urbanek, Williams, Agrawal, Romero-Soriano, and Drozdzal}]{manas2024texttoimage}
Oscar Mañas, Pietro Astolfi, Melissa Hall, Candace Ross, Jack Urbanek, Adina Williams, Aishwarya Agrawal, Adriana Romero-Soriano, and Michal Drozdzal. 2024.
\newblock \href {https://arxiv.org/abs/2403.17804} {Improving text-to-image consistency via automatic prompt optimization}.
\newblock \emph{Preprint}, arXiv:2403.17804.

\bibitem[{Ning et~al.(2021)Ning, Dong, Shi, Li, and Li}]{super_resolution}
Qian Ning, Weisheng Dong, Guangming Shi, Leida Li, and Xin Li. 2021.
\newblock \href {https://doi.org/10.1109/JSTSP.2020.3037516} {Accurate and lightweight image super-resolution with model-guided deep unfolding network}.
\newblock \emph{IEEE Journal of Selected Topics in Signal Processing}, 15(2):240--252.

\bibitem[{Podell et~al.(2023)Podell, English, Lacey, Blattmann, Dockhorn, Müller, Penna, and Rombach}]{podell2023sdxl}
Dustin Podell, Zion English, Kyle Lacey, Andreas Blattmann, Tim Dockhorn, Jonas Müller, Joe Penna, and Robin Rombach. 2023.
\newblock \href {https://arxiv.org/abs/2307.01952} {Sdxl: Improving latent diffusion models for high-resolution image synthesis}.
\newblock \emph{Preprint}, arXiv:2307.01952.

\bibitem[{Ruiz et~al.(2023)Ruiz, Li, Jampani, Pritch, Rubinstein, and Aberman}]{ruiz2023dreambooth}
Nataniel Ruiz, Yuanzhen Li, Varun Jampani, Yael Pritch, Michael Rubinstein, and Kfir Aberman. 2023.
\newblock Dreambooth: Fine tuning text-to-image diffusion models for subject-driven generation.
\newblock In \emph{Proceedings of the IEEE/CVF Conference on Computer Vision and Pattern Recognition}, pages 22500--22510.

\bibitem[{Shinn et~al.(2023)Shinn, Cassano, Gopinath, Narasimhan, and Yao}]{shinn2023reflexion}
Noah Shinn, Federico Cassano, Ashwin Gopinath, Karthik~R Narasimhan, and Shunyu Yao. 2023.
\newblock \href {https://openreview.net/forum?id=vAElhFcKW6} {Reflexion: language agents with verbal reinforcement learning}.
\newblock In \emph{Thirty-seventh Conference on Neural Information Processing Systems}.

\bibitem[{Sobania et~al.(2024)Sobania, Briesch, and Rothlauf}]{sobania2024comfygi}
Dominik Sobania, Martin Briesch, and Franz Rothlauf. 2024.
\newblock \href {https://arxiv.org/abs/2411.14193} {Comfygi: Automatic improvement of image generation workflows}.
\newblock \emph{Preprint}, arXiv:2411.14193.

\bibitem[{Wang et~al.(2025)Wang, Yang, Wang, Xu, Wang, Wang, Luo, Zhang, Hu, and Zhang}]{wang2025cigeval}
Jifang Wang, Xue Yang, Longyue Wang, Zhenran Xu, Yiyu Wang, Yaowei Wang, Weihua Luo, Kaifu Zhang, Baotian Hu, and Min Zhang. 2025.
\newblock \href {https://arxiv.org/abs/2504.07046} {A unified agentic framework for evaluating conditional image generation}.
\newblock \emph{Preprint}, arXiv:2504.07046.

\bibitem[{Wang et~al.(2024)Wang, Mao, Fried, and Neubig}]{wang2024agentworkflowmemory}
Zora~Zhiruo Wang, Jiayuan Mao, Daniel Fried, and Graham Neubig. 2024.
\newblock \href {https://arxiv.org/abs/2409.07429} {Agent workflow memory}.
\newblock \emph{Preprint}, arXiv:2409.07429.

\bibitem[{Wei et~al.(2022)Wei, Tay, Bommasani, Raffel, Zoph, Borgeaud, Yogatama, Bosma, Zhou, Metzler, Chi, Hashimoto, Vinyals, Liang, Dean, and Fedus}]{wei2022emergent}
Jason Wei, Yi~Tay, Rishi Bommasani, Colin Raffel, Barret Zoph, Sebastian Borgeaud, Dani Yogatama, Maarten Bosma, Denny Zhou, Donald Metzler, Ed~H. Chi, Tatsunori Hashimoto, Oriol Vinyals, Percy Liang, Jeff Dean, and William Fedus. 2022.
\newblock \href {https://openreview.net/forum?id=yzkSU5zdwD} {Emergent abilities of large language models}.
\newblock \emph{Transactions on Machine Learning Research}.
\newblock Survey Certification.

\bibitem[{Xu et~al.(2023)Xu, Shi, Hu, Yu, Li, Zhang, and Wu}]{xu2023reasoning}
Zhenran Xu, Senbao Shi, Baotian Hu, Jindi Yu, Dongfang Li, Min Zhang, and Yuxiang Wu. 2023.
\newblock \href {https://arxiv.org/abs/2311.08152} {Towards reasoning in large language models via multi-agent peer review collaboration}.
\newblock \emph{Preprint}, arXiv:2311.08152.

\bibitem[{Xu et~al.(2025)Xu, Wang, Hu, Wang, and Zhang}]{xu-etal-2025-mekb}
Zhenran Xu, Jifang Wang, Baotian Hu, Longyue Wang, and Min Zhang. 2025.
\newblock \href {https://aclanthology.org/2025.naacl-demo.33/} {{M}e{KB}-sim: Personal knowledge base-powered multi-agent simulation}.
\newblock In \emph{Proceedings of the 2025 Conference of the Nations of the Americas Chapter of the Association for Computational Linguistics: Human Language Technologies (System Demonstrations)}, pages 393--403, Albuquerque, New Mexico. Association for Computational Linguistics.

\bibitem[{Xue et~al.(2024)Xue, Lu, Huang, Wang, Ouyang, and Bai}]{xue2024comfybench}
Xiangyuan Xue, Zeyu Lu, Di~Huang, Zidong Wang, Wanli Ouyang, and Lei Bai. 2024.
\newblock \href {https://arxiv.org/abs/2409.01392} {Comfybench: Benchmarking llm-based agents in comfyui for autonomously designing collaborative ai systems}.
\newblock \emph{Preprint}, arXiv:2409.01392.

\bibitem[{Yang et~al.(2025)Yang, Liu, Chaudhary, Fakoor, Chaudhari, Karypis, and Rangwala}]{yang2025agentoccam}
Ke~Yang, Yao Liu, Sapana Chaudhary, Rasool Fakoor, Pratik Chaudhari, George Karypis, and Huzefa Rangwala. 2025.
\newblock \href {https://openreview.net/forum?id=oWdzUpOlkX} {Agentoccam: A simple yet strong baseline for {LLM}-based web agents}.
\newblock In \emph{The Thirteenth International Conference on Learning Representations}.

\bibitem[{Ye et~al.(2023)Ye, Zhang, Liu, Han, and Yang}]{ye2023ip}
Hu~Ye, Jun Zhang, Sibo Liu, Xiao Han, and Wei Yang. 2023.
\newblock Ip-adapter: Text compatible image prompt adapter for text-to-image diffusion models.
\newblock \emph{arXiv preprint arXiv:2308.06721}.

\bibitem[{Yin et~al.(2024)Yin, Brahman, Ravichander, Chandu, Chang, Choi, and Lin}]{yin-etal-2024-agent}
Da~Yin, Faeze Brahman, Abhilasha Ravichander, Khyathi Chandu, Kai-Wei Chang, Yejin Choi, and Bill~Yuchen Lin. 2024.
\newblock \href {https://doi.org/10.18653/v1/2024.acl-long.670} {Agent lumos: Unified and modular training for open-source language agents}.
\newblock In \emph{Proceedings of the 62nd Annual Meeting of the Association for Computational Linguistics (Volume 1: Long Papers)}, pages 12380--12403, Bangkok, Thailand. Association for Computational Linguistics.

\bibitem[{Yu et~al.(2024)Yu, Gu, Li, Hu, Kong, Wang, He, Qiao, and Dong}]{yu2024scaling}
Fanghua Yu, Jinjin Gu, Zheyuan Li, Jinfan Hu, Xiangtao Kong, Xintao Wang, Jingwen He, Yu~Qiao, and Chao Dong. 2024.
\newblock \href {https://arxiv.org/abs/2401.13627} {Scaling up to excellence: Practicing model scaling for photo-realistic image restoration in the wild}.
\newblock \emph{Preprint}, arXiv:2401.13627.

\bibitem[{Zeng et~al.(2024)Zeng, Liu, Lu, Wang, Liu, Dong, and Tang}]{zeng-etal-2024-agenttuning}
Aohan Zeng, Mingdao Liu, Rui Lu, Bowen Wang, Xiao Liu, Yuxiao Dong, and Jie Tang. 2024.
\newblock \href {https://doi.org/10.18653/v1/2024.findings-acl.181} {{A}gent{T}uning: Enabling generalized agent abilities for {LLM}s}.
\newblock In \emph{Findings of the Association for Computational Linguistics: ACL 2024}, pages 3053--3077, Bangkok, Thailand. Association for Computational Linguistics.

\bibitem[{Zhang et~al.(2023)Zhang, Rao, and Agrawala}]{zhang2023adding}
Lvmin Zhang, Anyi Rao, and Maneesh Agrawala. 2023.
\newblock Adding conditional control to text-to-image diffusion models.
\newblock In \emph{CVPR}.

\bibitem[{Zheng et~al.(2024)Zheng, Zhang, Zhang, Ye, and Luo}]{zheng-etal-2024-llamafactory}
Yaowei Zheng, Richong Zhang, Junhao Zhang, Yanhan Ye, and Zheyan Luo. 2024.
\newblock \href {https://doi.org/10.18653/v1/2024.acl-demos.38} {{L}lama{F}actory: Unified efficient fine-tuning of 100+ language models}.
\newblock In \emph{Proceedings of the 62nd Annual Meeting of the Association for Computational Linguistics (Volume 3: System Demonstrations)}, pages 400--410, Bangkok, Thailand. Association for Computational Linguistics.

\end{thebibliography}
